\documentclass{article}
\usepackage{spconf,amsmath,graphicx}
\usepackage{cite}
\usepackage{adjustbox}
\usepackage{multirow}
\usepackage{float}
\usepackage{booktabs}
\usepackage{array}
\usepackage{url}
\usepackage[dvipsnames]{xcolor}
\newcolumntype{+}{>{\global\let\currentrowstyle\relax}}
\newcolumntype{^}{>{\currentrowstyle}}
\newcommand{\rowstyle}[1]{\gdef\currentrowstyle{#1}%
  #1\ignorespaces
}

\graphicspath{{figures/}}

\title{VULNERABILITY ANALYSIS OF FACE MORPHING ATTACKS FROM LANDMARKS AND GENERATIVE ADVERSARIAL NETWORKS}
\name{Eklavya Sarkar, Pavel Korshunov, Laurent Colbois, S\'ebastien Marcel}
\address{Idiap Research Institute, Martigny, Switzerland\\ 
\begin{adjustbox}{width=\linewidth,center}
\texttt{\{eklavya.sarkar, pavel.korshunov, laurent.colbois, sebastien.marcel\}@idiap.ch}
\end{adjustbox}}

\begin{document}
%
\maketitle
\begin{abstract}
Morphing attacks is a threat to biometric systems where the biometric reference in an identity document can be altered. This form of attack presents an important issue in applications relying on identity documents such as border security or access control. Research in face morphing attack detection is developing rapidly, however very few datasets with several forms of attacks are publicly available. This paper bridges this gap by providing a new dataset with four different types of morphing attacks, based on OpenCV, FaceMorpher, WebMorph and a generative adversarial network (StyleGAN), generated with original face images from three public face datasets. We also conduct extensive experiments to assess the vulnerability of the state-of-the-art face recognition systems, notably FaceNet, VGG-Face, and ArcFace. The experiments demonstrate that VGG-Face, while being less accurate face recognition system compared to FaceNet, is also less vulnerable to morphing attacks. Also, we observed that na\"ive morphs generated with a StyleGAN do not pose a significant threat.
\end{abstract}
\begin{keywords}
Biometrics, Face Recognition, Vulnerability Analysis, Morphing Attack, StyleGAN 2
\end{keywords}
\section{INTRODUCTION}
\label{sec:introduction}
\begin{figure*}[ht]
\begin{minipage}[b]{\linewidth}
  \centering
  \centerline{\includegraphics[width=\textwidth]{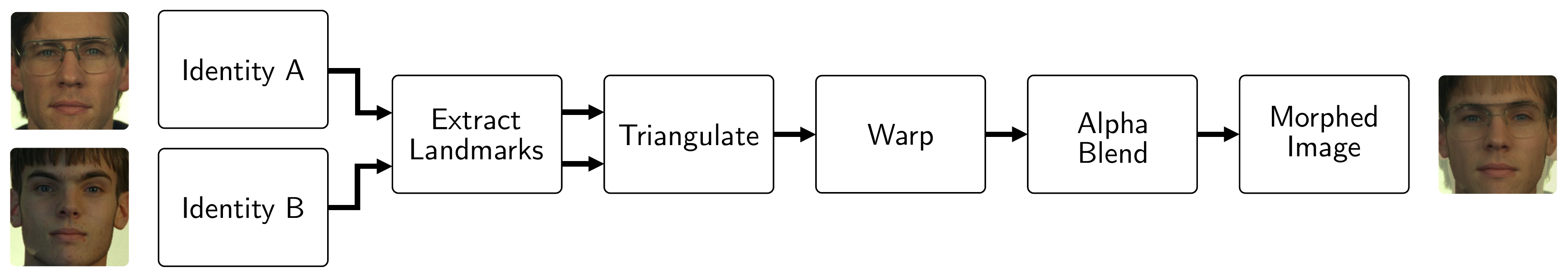}}
\end{minipage}
\caption{Morphed image generation pipeline for landmark-based methods.}
\label{fig:block}
\end{figure*}

After Ferrara \emph{et al.}~\cite{Ferrara2014} showed that by using a morphed photo of two different people an adversary can circumvent passport registration process, morphing attacks and how to detect them received a lot of attention from academic, industrial, and security communities. The vulnerability of state-of-the-art (SOTA) face recognition systems (FR) and the threat such vulnerability poses to the security systems relying on recognition technologies led to the explosion of research work in this area.

Most of the work related to morphing attacks (MAs) focuses on their detection. Recently proposed techniques for morphing attack detection (MAD) include methods based on \emph{so called} classical approaches using local binary patterns (LBP) and support vector machines (SVM)~\cite{Spreeuwers2018}, approaches rooted in image forensics that rely on photo response non uniformity (PRNU) function~\cite{Scherhag2019prnu}, deep neural networks specifically trained to detect morph images~\cite{Seibold2020}, and FR systems themselves serving as feature extractors for an support vector machine (SVM) classifier~\cite{Wandzik2018}. The National Institute of Standards and Technology (NIST) is now conducting independent evaluations of MAD technologies~\cite{NIST2020}.

Because the researchers are mainly focused on detecting MAs, some related questions suffer from lack of attention: i) very few public databases of morphed images are available and ii) little is reported on whether the latest SOTA face recognition systems remain vulnerable to the typical morphing attacks.

It is common for MAD systems to be evaluated on morphed images that are privately generated using real data from public datasets~\cite{Damer2018,Wandzik2018,deep_face_representations} and open source tools for face morphing such as OpenCV~\cite{opencv}. Very few databases are publicly available with a few exceptions, notably, the Advanced Multimedia Security Lab's (AMSL) Face Morph Image dataset~\cite{StirTrace}, which is the exception that proves the common practice. Also, the advent of generative adversarial networks (GANs) opens up new possibilities for generating possibly more realistic \emph{deep morphing} attacks~\cite{korshunov2019deepmorphs}, which are not yet well explored.

Therefore, in this paper, we aim to fill these gaps by proposing several publicly available datasets with morphed images, including with the latest GAN-based morphs, and an extensive vulnerability assessment of the SOTA face recognition systems. Using open source morphing implementations OpenCV-based~\cite{opencv} and FaceMorpher~\cite{facemorpher_repo}, online tool WebMorph~\cite{webmorph}, and the latest StyleGAN 2~\cite{stylegan2}, we generated morphed images using real faces from publicly available FERET~\cite{feret} and FRGCv2~\cite{frgc} datasets and also extend AMSL Face Morph Image~\cite{StirTrace} dataset. 

We use these three different datasets and different types of morphs to assess the vulnerability of SOTA systems such as FaceNet~\cite{facenet} and VGG-Face~\cite{deepfacerecognition}, as well as ArcFace~\cite{arcface_paper}, which is used in some of the latest morphing attacks vulnerability studies~\cite{deep_face_representations}, and baseline ``classical'' systems based on Gabor filters~\cite{gabor} and Inter-Session Variability (ISV) modelling~\cite{isv}. Therefore, besides providing several morphed images to public, this paper is also an important milestone in our understanding of where the current state of the art morph generation algorithms and FR systems are at.

To allow researchers to verify, reproduce, and extend our work, we provide all scripts for generation of morphed images for FERET, FRGC, and FRLL (the dataset ASML Face Morph Image set is based on) datasets, FR systems used, and the scripts for their vulnerability assessment as open source\footnote{\url{https://gitlab.idiap.ch/bob/bob.paper.icassp2021_morph}}. Please note that due to the licensing restrictions of the original FERET, FRGC, and FRLL datasets, we are unable to provide the morphed images directly, but we do provide easily installable and well documented code that can be used to re-generate all the morphs we created, given one has legally obtained the copies of the original image datasets.

\section{MORPH GENERATION}
\label{sec:morph_generation}

In this section, we present the datasets with bona fide faces and the different tools, including the one based on GANs, used to generate the morphing images for the vulnerability experiments.

\begin{figure*}[ht]
\begin{minipage}[b]{0.139\linewidth}
  \centering
  \centerline{\includegraphics[width=\textwidth]{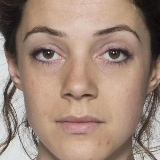}}
  \centerline{(a) Identity A}\medskip
\end{minipage}
\begin{minipage}[b]{0.139\linewidth}
  \centering
  \centerline{\includegraphics[width=\textwidth]{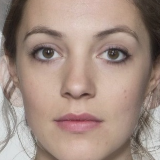}}
  \centerline{(b) OpenCV}\medskip
\end{minipage}
\begin{minipage}[b]{0.139\linewidth}
  \centering
  \centerline{\includegraphics[width=\textwidth]{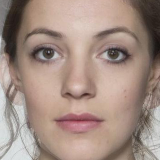}}
  \centerline{(c) FaceMorpher}\medskip
\end{minipage}
\begin{minipage}[b]{0.139\linewidth}
  \centering
  \centerline{\includegraphics[width=\textwidth]{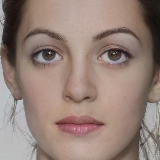}}
  \centerline{(d) StyleGAN2}\medskip
\end{minipage}
\begin{minipage}[b]{0.139\linewidth}
  \centering
  \centerline{\includegraphics[width=\textwidth]{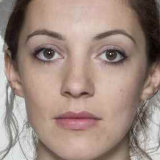}}
  \centerline{(e) WebMorph}\medskip
\end{minipage}
\begin{minipage}[b]{0.139\linewidth}
  \centering
  \centerline{\includegraphics[width=\textwidth]{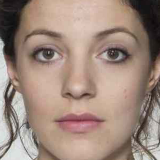}}
  \centerline{(f) CombinedMorph}\medskip
\end{minipage}
\begin{minipage}[b]{0.139\linewidth}
  \centering
  \centerline{\includegraphics[width=\textwidth]{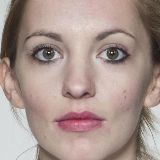}}
  \centerline{(g) Identity B}\medskip
\end{minipage}
\caption{Different types of generated morphed images on the FRLL dataset after initial preprocessing.}
\label{fig:morph_figures}
\end{figure*}

\subsection{Datasets}
\label{ssec:database}
We used FERET~\cite{feret}, FRGC v2.0~\cite{frgc}, and  Face Research Lab London (FRLL)~\cite{london} datasets of facial images to generate the morphs. FERET and FRGC were selected because they are \emph{de facto} the standard datasets commonly used in papers on morphing attack detection~\cite{pad,deep_face_representations} and they have large number of images of different identities. We also used FRLL dataset, because the only practically available dataset of morphed images, AMSL Face Morph Image dataset generated by Neubert \emph{et al.}~\cite{StirTrace} used the original faces from FRLL dataset. 
Also, FRLL dataset is a great choice to use for creating morphing attacks, because it contains close-up frontal face images of very high visual quality and $1350 \times 1350$ resolution, shot under \textit{uniform} illumination with large varieties in ethnicity, pose, and expression. Each face is annotated using $189$ facial landmarks, which is notably a very high number, as typical landmarks detectors provide no more than $68$-$70$ landmarks. The main limitation of FRLL dataset, compared to FERET and FRGC datasets, is the limited number $102$ of different identities with $53$ males and $49$ females. 

For each dataset, we select bona fide (or original) face pairs for morph generation by following the existing protocols used in previous work. For FERET and FRGC, we follow the protocols used in the work by Scherhag \emph{et al.}~\cite{deep_face_representations} that were kindly provided by the authors (though they were not able to provide any morphed images). For FRLL dataset, we follow the protocols used in AMSL Face Morph Image dataset by Neubert \emph{et al.}~\cite{StirTrace}. Using these protocols (essentially, which facial image pairs to morph), we generated morphs with the following morphing tools: OpenCV, FaceMorpher, WebMorph, and StyleGAN 2, which are described in Section~\ref{ssec:morphing_tools}.

\subsection{Morphing Tools}
\label{ssec:morphing_tools}
As morphing tools, we selected two commonly used open source face morphing algorithms, OpenCV-based~\cite{opencv} and FaceMorpher~\cite{facemorpher_repo}, web-based open source morphing tool called WebMorph~\cite{webmorph}, and the algorithm adapted from the recently proposed StyleGAN 2 implementation~\cite{stylegan2}.


%

\subsubsection{Landmarks-Based Morphs}

The \textbf{OpenCV} algorithm is an adaptation of an open-source implementation~\cite{opencv} used for morphing faces using $68$-point annotator from Dlib library~\cite{dlib09}. Face landmarks are obtained for each of the two bona fide source images and are used to form Delaunay triangles, which are in-turn warped and alpha blended. 

\textbf{FaceMorpher} is also an open-source~\cite{facemorpher_repo} similar to OpenCV landmark-based morphing algorithm, but with the STASM~\cite{stasm} landmark detector instead. Both algorithms create morphs with noticeable ghosting artefacts for all three datasets, as the region outside the area covered by these landmarks is simply averaged.

\textbf{WebMorph}~\cite{webmorph} is an online landmark-based morphing tool created by the FRLL dataset providers, which requires $189$ landmarks that are available only in FLLR dataset, to generate morphed images with better alignment and of an overall higher visual quality. Ghosting artefacts are still visible and prominent around the hair and neck area, but are noticeably improved around the ears. As this tool works exclusively with the annotation files of FRLL dataset, we were not able to generate the same types of morphs for FERET and FRGC datasets.

The morphed images in the AMSL dataset, which were generated by Neubert \emph{et al.}~\cite{StirTrace} from bona fide images of FRLL dataset using the private Combined Morphs tool, contain very realistic morphs with virtually no ghosting artefacts, even around the hair and neck area, because of the additional poisson image editing. Unlike the other tools, this proprietary technique generates \textit{two} unique morphed images for every pair of source bona fide images. Since Combined Morphs is a  proprietary tool, we were unable use it for the generation of the same morphs for the FERET and FRGC datasets.


\subsubsection{Generative Adversarial Network-Based Morphs}
Following the advances in generative adversarial networks (GANs), there were attempts to generate morphed images using a GAN instead of landmark-based methods~\cite{GAN_Morphs_Threat,korshunov2019deepmorphs}. In this paper, we adapted the latest~\textbf{StyleGAN 2}~\cite{stylegan2} to develop a morphing algorithm which can generate high resolution realistic looking faces with no noticeable artifacts. The StyleGAN 2 was pre-trained on the FFHQ dataset introduced in~\cite{stylegan}. 

The faces are cropped to obtain the same landmark alignment as in the FFHQ dataset. The images are then projected into the $W$ space of StyleGAN 2 by optimizing the input latent style vector that is fed to the generator network, such that it minimizes the perceptual loss between the generated and real image~\cite{stylegan2}. Once an associated latent vector has been computed for each of the source images, morphs can be generated by linearly interpolating between two latent vectors, and feeding the interpolated vector back into the generator.

This technique yields very realistic looking morphs without visual artefacts, however, since StyleGAN does not have any information about the identities in bona fide images, there is no guarantee that the resulted morph is actually a blen of these identities (see the example in Figure~\ref{fig:morph_figures}(d) for an idea). 

StyleGAN 2 also requires the projected images to be at a high resolution (1024x1024 after cropping), and works better with an uniform background, which makes the FRLL dataset particularly appropriate. A side not observation of using StyleGAN 2 for generating morphs is that it is equally easy to generate high-quality morphs for smiling expressions as it is for the neutral faces, which is not possible with typical landmark-based tools.

\subsection{Generation}
\label{ssec:generation}
Using morphing tools presented in Section~\ref{ssec:morphing_tools}, we generate three sets of morphs each consisting of $1'222$, $529$, and $964$ images for FRLL, FERET, and FRGC datasets respectively. For FRLL, we also generate one additional set of morphs using WebMorph tool. Please note that the morphs for FERET and FRGC datasets are generated using the same protocols used in~\cite{deep_face_representations}, while the morph generation protocol defined in ASML Face Morph Image dataset was used in case of FRLL.


\section{Evaluation Protocol}
\label{sec:evaluation_protocol}

\subsection{Face Recognition Systems}
\label{ssec:frs_pipelines}

To evaluate vulnerability of face recognition against morphing attacks, we used publicly available pre-trained FaceNet, ArcFace, and VGG-Face  architectures. We used the last fully connected layers of these networks as features and the cosine distance as a classifier. For a given test face, the confidence score of whether it belongs to a reference model is the cosine distance between the average reference feature vector and the feature vector of a test face. These systems are the state of the art recognition systems with Facenet showing $99.63\%$~\cite{facenet}, ArcFace -- $99.53$~\cite{arcface_paper}, and VGG-Face -- $98.95\%$~\cite{deepfacerecognition} accuracies on the labeled faces in the wild (LFW) dataset.

We also used two ``classical'' baselines: i) Gabor jet implementation from LBP features~\cite{gabor} and ii) ISV-based face recognition~\cite{isv}. DCT features computed on overlapping blocks of 40x40 were used for the ISV-based system of $512$ Gaussian mixture models (GMMs) and $160$ dimensional subspace, which was pre-trained on the MOBIO~\cite{mobio} dataset.

\subsection{Evaluation Metrics}
\label{ssec:evaluation_metrics}
In a the verification process, a user attempting to authenticate presents a biometric probe and a claimed identity, and can be classified into one of the following three categories. \textit{A) Genuine user} (BF): probe and claimed identity both correctly belong to the user. \textit{B) Zero-effort impostor} (BF): probe belongs to the user, but the claimed identity corresponds to a different enrolled user. \textit{C) Morph attack impostor} (MA): probe matches the claimed identity but does not correspond to the user.

The \textit{verification} performance is typically evaluated with the following metrics. \textit{A) False Match Rate (FMR)}~\cite{pad}: proportion of zero-effort impostors that are falsely authenticated. \textit{B) False Non-Match Rate (FNMR)}~\cite{pad}: proportion of genuine users which are falsely rejected. \textit{C) Mated Morph Presentation Match Rate  (MMPMR)}~\cite{Biometric_Systems_under_Morphing_Attacks}: proportion of morphs attacks impostors accepted by the face recognition system.


\subsection{Evaluation scenarios}
\label{ssec:scenarios}
For the FERET and FRGC datasets, we adopted the same evaluation scenarios used in~\cite{deep_face_representations}. For the FRLL dataset, we defined our own evaluation protocols due to the lack of publicly available protocols for FRLL. 

In general, there are two main scenarios under which a face recognition system is evaluated: a bona fide (BF) scenario where both the reference and probes images as genuine, so there are no attacks and the system is assumed to perform under the conditions it was designed for; and the morphing attack (MA) scenario when morphs are introduced to the face recognition with a malicious intent to spoof the recognition. There are also two variants of MA scenario, when a morphed image can be either used as a reference, i.e., FR system is hijacked during enrollment process (typical morphing attack scenario), or a morphed image is used as a probe, which is similar to presentation attack scenario. The number of reference and probe images for each evaluation scenario is summarized in Table~\ref{table:comparaisons}.

Also note that we did not split datasets into training, development, and test subsets but used each whole dataset one test set. The reason for this is that all recognition systems we used were pre-trained on other databases, so there is no need for the training set, and we choose the decision threshold to compute MMPMR value for MA scenario based on FMR value computed in the bona fide scenario, so there is no need for a development set.


\begin{table}[ht]
    \centering
    \caption{Number of images in different evaluation scenarios.}
    \label{table:comparaisons}
    \begin{tabular}{+l^l^l^l^l^l}
    \toprule
    \rowstyle{\bfseries} Dataset & Morphs as & BF & MA & Impostors \\
    \midrule
    \multirow{2}{*}{FRLL}  & References & 91  & 584 & 1,984 \\ 
                           & Probes     & 584 & 91  & 4,153 \\ 
    \midrule
    \multirow{2}{*}{FERET} & References & 529 & 791 & 418,439 \\ 
                           & Probes     & 791 & 529 & 418,439 \\ 
    \midrule
    \multirow{2}{*}{FRGC} & References  & 3,298 & 964 & 1,698,384 \\ 
                          & Probes      & 964 & 3,298 & 1,698,384 \\ 
    \bottomrule
    \end{tabular}
\end{table}

\section{EXPERIMENTAL RESULTS}
\label{sec:experimental_results}
Table~\ref{table:all_iamprs} summarizes the results of vulnerability assessment of the face recognition systems described in Section~\ref{ssec:frs_pipelines} under different morphing attack scenarios (see Section~\ref{ssec:scenarios} for details). The MMPMR metric (see Section~\ref{ssec:evaluation_metrics} for details) is calculated by setting the decision threshold at FMR=$0.1$\% in the bona fide scenario.
\begin{table}[h]
    \centering
    \caption{MMPMR @ FMR = 0.1\% (morphs as references | morphs as probes) [\%]}
    \label{table:all_iamprs}
    \begin{adjustbox}{width=\columnwidth,center}
    \begin{tabular}{+l^l^c^c^c^c^c^c}
    \toprule
    \rowstyle{\bfseries} Dataset & FRS & OpenCV & FaceMorpher & StyleGAN2 & WebMorph & AMSL \\ 
    \midrule
    \multirow{5}{*}{FRLL}   & FaceNet & 83.3 | 72.0  & 64.5 | 68.2 & 5.9  | 11.0 & 82.7 | 70.8 & 89.2  | 92.5 \\
                            & ArcFace & 59.8 | 73.8  & 57.6 | 75.3 & 9.8  | 18.3 & 60.9 | 73.8 & 58.0  | 79.4 \\
                            & VGG     & 39.7 | 48.6  & 23.4 | 47.1 & 3.0  | 9.1 & 38.2  | 52.2 & 65.7  | 89.8 \\
                            & Gabor   & 87.2 | 100.0 & 83.9 | 99.4 & 11.8 | 37.9 & 85.4 | 100.0 & 86.3 | 99.9 \\
                            & ISV     & 59.8 | 97.8  & 56.1 | 96.1 & 9.2  | 43.6 & 59.5 | 97.4 & 55.3  | 99.9 \\
    \midrule
    \multirow{5}{*}{FERET} & FaceNet & 41.1 | 40.6 & 39.9 | 40.3 & 1.6 | 1.3  & N/A & N/A \\
                           & Arcface & 34.6 | 35.2 & 34.1 | 34.8 & 2.4 | 2.5  & N/A & N/A \\
                           & VGG     & 22.0 | 21.0 & 20.5 | 18.3 & 2.0 | 1.5  & N/A & N/A \\
                           & Gabor   & 66.6 | 90.9 & 63.7 | 88.5 & 1.3 | 40.8 & N/A & N/A \\
                           & ISV     & 44.8 | 58.4 & 42.6 | 56.5 & 2.7 | 3.4  & N/A & N/A \\
    \midrule
    \multirow{5}{*}{FRGC}  & FaceNet & 6.9  | 5.9    & 7.0  | 5.7   & 1.0 | 0.7  & N/A & N/A \\
                           & Arcface & 11.9 | 10.8   & 12.1 | 11.2  & 0.5 | 0.4  & N/A & N/A \\
                           & VGG     & 5.5  | 4.5    & 5.1  | 4.8   & 0.7 | 0.4  & N/A & N/A \\
                           & Gabor   & 7.1  | 80.8   & 6.7  | 81.0  & 0.6 | 75.8 & N/A & N/A \\
                           & ISV     & 4.2  | 6.5    & 3.5  | 6.2   & 0.6 | 0.6  & N/A & N/A \\
    \bottomrule
    \end{tabular}
    \end{adjustbox}
\end{table}

The results in Table~\ref{table:all_iamprs} illustrate two important observations: 
\begin{enumerate}
    \item The more accurate face recognition systems are the more vulnerable they are to the morphing attacks regardless of whether they are used as references or as probes, which is also in line with the observations made for presentation attacks~\cite{deeply_vulnerable}. This trend is especially evident when we compare a more accurate and deeper FaceNet architecture with VGG-Face for all databases and types of morphs.
    \item The morphs generated with  StyleGAN 2, one of the latest generative adversarial network, do not pose significant threats to the state of the art recognition systems. A slight elevation in the MMPMR error for StyleGAN2 morphs of FRLL dataset indicates that high quality original images may lead to slightly more accurate morphs.  
\end{enumerate}




\section{CONCLUSION}
\label{sec:conclusion}

The paper presents an extensive vulnerability assessment of the state of the art face recognition systems based on VGG-Face, ArcFace, and FaceNet neural network models on three image databases with five different morphing attacks, one of which was created using StyleGAN 2. The experiments demonstrate that a more accurate face recognition system FaceNet is more vulnerable to the morphing attacks and that GAN-based morph do not yet pose a significant threat to modern recognition systems.
However, if one would introduce an identity loss into StyleGAN-based morph generation to ensure that the identities of both bona fide inputs are preserved in the resulted morph, then the GAN-based morphs may become highly threatening to face recognition and this would be an interesting future direction. 

\vfill
\pagebreak
\bibliographystyle{IEEEbib}
{\footnotesize \bibliography{refs}} 

\end{document}